\documentclass[sigconf]{acmart}

\copyrightyear{2025}
\acmYear{2025}
\setcopyright{acmlicensed}
\acmConference[KDD '25] {Proceedings of the 31st ACM SIGKDD Conference on Knowledge Discovery and Data Mining V.1}{August 3--7, 2025}{Toronto, ON, Canada.}
\acmBooktitle{Proceedings of the 31st ACM SIGKDD Conference on Knowledge Discovery and Data Mining V.1 (KDD '25), August 3--7, 2025, Toronto, ON, Canada}
\acmISBN{979-8-4007-1245-6/25/08}
\acmDOI{10.1145/3690624.3709391}

\usepackage{algorithm}
\usepackage{algorithmic}

\usepackage[utf8]{inputenc} 
\usepackage[T1]{fontenc}    
\usepackage{url}            
\usepackage{booktabs}       
\usepackage{amsfonts}       
\usepackage{nicefrac}       
\usepackage{microtype}      
\usepackage{amsthm}
\usepackage{amsmath}
\usepackage{graphicx}

\usepackage{bbm}
\usepackage{diagbox}
\usepackage{colortbl}
\usepackage{multirow}
\usepackage{pifont} 
\usepackage{bm}
\usepackage{subcaption}
\usepackage{color,xcolor}
\usepackage[font=small,labelfont=bf]{caption}  
\usepackage{makecell}
\usepackage{tabulary}
\definecolor{demphcolor}{RGB}{144,144,144}
\newcommand{\demph}[1]{\textcolor{demphcolor}{#1}}
\definecolor{mygray}{gray}{0.4}
\usepackage{colortbl}
\usepackage{enumitem}
\usepackage{multirow}
\usepackage{threeparttable}
\usepackage{multirow}
\usepackage[export]{adjustbox}
\usepackage{algorithm}
\usepackage{algorithmic}
\usepackage{mdframed}

\newcommand{\tabincell}[2]{\begin{tabular}{@{}#1@{}}#2\end{tabular}}

\usepackage{makecell}
\usepackage{xcolor} 
\definecolor{snowblue}{rgb}{0.65, 0.80, 0.98} 

\usepackage{newfloat}
\usepackage{listings}
\usepackage{balance}
\AtBeginDocument{%
  }

\AtBeginDocument{%
  }

\begin{document}

\title{DynST: Dynamic Sparse Training for Resource-Constrained Spatio-Temporal Forecasting}









\author{Hao Wu}
\authornote{This work was completed during an internship in the Machine Learning Platform Department at Tencent.}
\authornote{Equal contribution}
\affiliation{%
  \institution{University of Science and Technology of China}
  \city{Hefei}
  \state{Anhui}
  \country{China}
}
\email{wuhao2022@mail.ustc.edu.cn}

\author{Haomin Wen}
\authornotemark[2]
\affiliation{%
  \institution{Carnegie Mellon University}
  \city{Pittsburgh}
  \state{Pennsylvania}
  \country{USA}
}
\email{wenhaomin.whm@gmail.com}

\author{Guibin Zhang}
\affiliation{%
  \institution{Tongji University}
  \city{Shanghai}
  \state{Shanghai}
  \country{China}
}
\email{bin2003@tongji.edu.cn}

\author{Yutong Xia}
\affiliation{%
  \institution{National University of Singapore}
  \city{Singapore}
  \state{Singapore}
  \country{Singapore}
}
\email{yutong.x@outlook.com}

\author{Yuxuan Liang}
\authornote{Corresponding author.}
\affiliation{%
  \institution{Hong Kong University of Science and Technology (Guangzhou)}
  \city{Guangzhou}
  \state{Guangdong}
  \country{China}
}
\email{yuxliang@outlook.com}

\author{Yu Zheng}
\affiliation{%
  \institution{JD iCity, JD Technology}
  \city{Beijing}
  \state{Beijing}
  \country{China}
}
\email{msyuzheng@outlook.com}

\author{Qingsong Wen}
\affiliation{%
  \institution{Squirrel AI}
  \city{Seattle}
  \state{Washington}
  \country{USA}
}
\email{qingsongedu@gmail.com}

\author{Kun Wang}
\authornotemark[3]
\affiliation{%
  \institution{Nanyang Technological University}
  \city{Singapore}
  \state{Singapore}
  \country{Singapore}
}
\email{wk520529@mail.ustc.edu.cn}
\renewcommand{\shortauthors}{Hao Wu et al.}


\begin{abstract}
  The ever-increasing sensor service, though opening a precious path and providing a deluge of earth system data for deep-learning-oriented earth science, sadly introduce a daunting obstacle to their industrial level deployment. Concretely, earth science systems rely heavily on the extensive deployment of sensors, however, the data collection from sensors is constrained by complex geographical and social factors, making it challenging to achieve comprehensive coverage and uniform deployment. To alleviate the obstacle, traditional approaches to sensor deployment utilize specific algorithms to design and deploy sensors. These methods \textit{dynamically adjust the activation times of sensors to optimize the detection process across each sub-region}. Regrettably, formulating an activation strategy generally based on historical observations and geographic characteristics, which make the methods and resultant models were neither simple nor practical. Worse still, the complex technical design may ultimately lead to a model with weak generalizability. In this paper, we introduce for the first time the concept of spatio-temporal data dynamic sparse training and are committed to adaptively, dynamically filtering important sensor distributions. To our knowledge, this is the \textbf{first} proposal (\textit{termed DynST}) of an \textbf{industry-level} deployment optimization concept at the data level. However, due to the existence of the temporal dimension, pruning of spatio-temporal data may lead to conflicts at different timestamps. To achieve this goal, we employ dynamic merge technology, along with ingenious dimensional mapping to mitigate potential impacts caused by the temporal aspect. During the training process, DynST utilize iterative pruning and sparse training, repeatedly identifying and dynamically removing sensor perception areas that contribute the least to future predictions.
  
  DynST demonstrates tremendous capability on industrial-grade data from \textbf{JD Technology} TaxiBJ+ and practical deployment scenarios such as meteorology, combustion dynamics, and turbulence. It seamlessly integrates with relevant models and efficiently prunes image and graph-type data, leading to significantly higher inference speeds without introducing noticeable performance degradation.
    
\end{abstract}





\maketitle

\section{Introduction}


Deep learning has revolutionized spatio-temporal (ST) forecasting, demonstrating remarkable proficiency in distilling valuable insights from extensive ST datasets (\textit{e.g.,} human mobility \cite{wu2023earthfarseer, pan2019urban}, precipitation \cite{zhang2023skilful,bi2023accurate}, frame dynamics \cite{li2020fourier,wu2023solving}, and meteorology \cite{pathak2022fourcastnet, wu2023pastnet}). In recent years, the widespread deployment of sensors has ushered in an unprecedented influx of earth system data from across the globe and outer space. However, this expansion comes at a significant cost. Worse still, the prolonged operation of sensors leads to significant power loss and hardware wear. To illustrate, the National Science Foundation (NSF) in the United States allocated over one billion dollars in its 2021 fiscal year budget to support research in these areas at numerous universities nationwide \cite{rissler2020gender}.

Traditional approaches to sensor deployment optimization \cite{priyadarshi2020wireless,zou2003sensor,yarinezhad2023sensor,xu2020optimal,kundu2023study}, such as virtual force and Voronoi diagrams, utilize specific algorithms to select the regions where sensors should be deployed. These methods \textit{ideally adjust the activation of sensors to optimize the detection process across each sub-region}.

\textbf{Research Gap.} Unfortunately, generating an effective activation strategy using only pre-existing historical observation data or urban geographic characteristics is very tricky, as it often involves complex technical design \cite{zhang2023demand}. Furthermore, with numerous factors influencing sensor deployment, relying solely on single variables (such as urban layout or geographic features) does not accurately capture the optimal deployment strategy \cite{yan2023survey,zheng2023spatial}.

With this in mind, in this paper, our aim is to \textbf{speedup inference and training time by proposing a novel sensor deactivation strategy, which is based on historical observations}. A promising direction and motivation involves adopting deep-learning-oriented metrics to adaptively and dynamically evaluate or verify the benefits brought by each sensor deployment. The ever-increasing dynamic sparse training (termed DST) \cite{evci2020rigging, liu2021we, huang2023dynamic,liu2020dynamic}, though opening a potential path for automating effective deployment, still remains in its nascent stages when exploring spatio-temporal scenarios. 

However, transferring the concept of Dynamic Sparse Training to the realm of spatio-temporal forecasting is intuitively beneficial as it can significantly accelerate model training while optimizing deployment. Specifically, DST technology shows promise in training a sub-network from scratch, employing sparse network training strategies, to achieve the performance levels of a fully dense network. In practical terms, if sensor deployment nodes are considered as intuitive data distribution collection points, both the training of models and the optimization of sensors remain computationally intensive tasks in both academic and industrial settings.

Regrettably, the application of DST to the challenge of spatio-temporal sensor deployment necessitates a meticulously aligned methodology. This is primarily because there exists a pronounced and inherent disparity between conventional DST frameworks and the nuances of spatio-temporal forecasting. Specifically: 

\begin{itemize}[leftmargin=*]
    \item[\ding{224}]  \textit{DST focuses primarily at the network level; if we abstract each sub-region of the data as the monitoring range of a sensor, DST methods struggle to dynamically select the most important sensors (or sub-counterpart of dataset) because the data is a pre-requisite and non-trainable.}

    \item[\ding{224}] \textit{The complexity of the above issue is further amplified in time-series data, where the spatial collection of information is dynamic. This dynamic nature poses a significant challenge in determining from historical data which elements will have a more substantial impact on future outcomes.}
\end{itemize}

\noindent To bridge the gap between industry and academia, this paper introduces for the first time the concept of dynamic sparse training for spatio-temporal data, termed DynST. \textbf{DynST dynamically trains to filter out the crucial parts of data for future predictions, and eliminates non-essential services to achieve resource-constrained service management.} Concretely, DynST utilizes dynamic training through a differentiable mask applied to historical regions, aiming to significantly reduce the proliferation of sensor deployment. This approach is taken at the algorithmic level to more effectively mask individual regions (each corresponding to a sensor device). Given the dynamic nature of time-series data, we utilize explicit \textbf{channel stacking} to construct overlapping saliency maps of historical regions. This facilitates the scoring of the importance of sensors in each region. DynST is both model-agnostic and efficient, demonstrating powerful optimization capabilities across a variety of industrial scenarios. It effectively reduces historically insignificant observation areas (\textit{i.e.}, sub-regions) in both regular and inherently irregular data environments, without impacting the performance of future predictions. 

\textbf{Summary of Contributions.} This paper makes multiple contributions to address the questions raised. Unlike the pruning of convolutional networks, which are typically heavily over-parameterized \cite{gao2022simvp, tan2022simvp,wang2018predrnn++, wang2019memory, gao2022earthformer, bai2022rainformer}, directly pruning a less parameterized spatio-temporal model offers limited scope for improvement. \textbf{Our first technical innovation} is the introduction of an end-to-end optimization framework called DynST, which uniquely prunes the sub-counterparts of data input for the first time. DynST does not rely on any specific spatio-temporal regular architecture or irregular graph structure \cite{scarselli2008graph, wu2020comprehensive}, allowing it to be flexibly applied across a wide range of spatio-temporal learning scenarios at scale. \textbf{To the best of our knowledge, this is the first work to employ dynamic sparse training techniques for the optimization of industrial-level devices.}

Viewing DynST as an advanced form of pruning for spatio-temporal datasets, our \textbf{second technical breakthrough} introduces a novel \underline{research direction.} This direction involves the utilization of deep-learning-guided sparse training techniques for the strategic optimization of sensor deployments. Our methodology is inherently adaptive and data-driven, focusing on identifying and preserving the most vital monitoring areas within historical data. This approach significantly diverges from traditional sensor deployment strategies~\cite{priyadarshi2020wireless, zou2003sensor, yarinezhad2023sensor, xu2020optimal, kundu2023study}, which often employ specific algorithmic designs for sensor placement, like virtual force techniques and Voronoi diagrams. \textbf{In contrast, our approach offers substantial real-world relevance and industrial applicability, representing a major leap forward in the field.}

Our proposal has been experimentally verified across various industrial-grade datasets and diverse backbones. The key observations from our study are outlined below:
\begin{itemize}[leftmargin=*]
    \item \textbf{DynST Maintains Performance in Sparse Data.} DynST integrates into various models and handles sparser input data without significantly affecting performance. For example, in the GNN architecture, DynST integration slightly increases the MAE on the Turbulence dataset from $4.35\rightarrow4.37$. In the Transformer architecture, DynST reduces the MAE from $3.67\rightarrow3.59$ on the JD traffic benchmark.

    \item \textbf{Significantly Improves Inference Efficiency.} DynST enhances inference speed across different architectures. On the Turbulence dataset, the STGCN architecture speeds up by 72\% to 1.721 times with DynST. On the Fire dataset, the GNN architecture speeds up by about 14.5\% to 1.541 times. On the JD Taxibj+ dataset, the Transform architecture nearly doubles in speed, increasing by about 34.5\% to 1.987 times. These examples demonstrate DynST's ability to improve computational efficiency, speed up inference, and handle large datasets efficiently.

    \item \textbf{Meets Industrial Standards.} DynST effectively meets industrial requirements, introducing minimal performance loss at sparsity levels ranging from $30\%\sim60\%$. Moreover, due to its model-agnostic nature, DynST is compatible with almost all industry-available models without conflict, showcasing strong transferability and plug-and-play characteristics.
    
\end{itemize}

\section{Related Work}

Our research is highly relevant to the following research themes:

\textbf{ST predictive learning} can be categorized into three main types. \textit{Convolutional Neural Network (CNN)-based architectures:} This research focuses on spatial feature extraction using CNN-based structures~\cite{gao2022simvp, tan2022simvp, wu2023pastnet, shi2015convolutional}. These architectures use convolutional layers to effectively detect patterns in image and video data. Key advancements include deep convolutional networks for complex feature extraction and 3D convolutions for spatial-temporal analysis in video processing~\cite{wang2018eidetic}; \textit{Recurrent Neural Network (RNN)-based Architectures:} RNNs are used to optimize temporal data handling~\cite{wang2017predrnn, wang2018predrnn++, wang2019memory},  which are key for tasks like sequence prediction and time-dependent data analysis; \textit{Transformer-based Architectures} delve into Transformer-based architectures for spatio-temporal data handling~\cite{gao2022earthformer, bai2022rainformer, wu2023earthfarseer, wu2023pastnet}, by employing their self-attention mechanism to effectively manage sequence data. They capture long-range dependencies in both spatial and temporal dimensions, making them suitable for complex sequence modeling and analysis. Notably, there are models that leverage graph neural networks primarily for ST graph management \cite{ji2023spatio, shao2022spatial, li2017diffusion}, which we will discuss later.

\textbf{Graph Neural Networks (GNNs) \& Graph Pooling.} GNNs have emerged as a prominent subfield in machine learning, specifically tailored to manage and analyze graph-structured data \cite{wang2022searching, yu2020representative, thekumparampil2018attention, you2019position}. In general, GNNs owe their efficacy to a distinct ``message-passing'' mechanism, which seamlessly integrates topological structures with node characteristics to yield richer graph representations. Leveraging the powerful topological awareness capabilities of GNNs, many studies have customized and adapted GNNs for predictions in spatio-temporal scenarios \cite{ji2023spatio, shao2022spatial, li2017diffusion}. Our method of dynamically filtering sensors can be understood as a form of graph pooling in the graph domain \cite{chen2018fastgcn, eden2018provable, chen2021unified, gao2019graph,ranjan2020asap, zhang2021hierarchical}. The distinction lies in the fact that traditional graph pooling is static, whereas our approach represents the first instance of addressing this kind of problem in dynamic temporal graphs.

\textbf{Senor Deployment.} In the field of sensor deployment, traditional methods \cite{priyadarshi2020wireless,zou2003sensor,yarinezhad2023sensor,xu2020optimal,kundu2023study} often employ specific algorithms, such as virtual force and Voronoi diagrams, for sensor design and deployment. These strategies involve dynamically adjusting sensor activation times to optimize detection across various sub-regions. However, developing an effective activation strategy based solely on historical observation data or urban geographic features presents significant challenges, primarily due to the intricate technical design requirements \cite{zhang2023demand}. Additionally, as highlighted in \cite{yan2023survey,zheng2023spatial}, focusing only on single variables like urban layout or geographic characteristics fails to fully address the complexities of optimal deployment strategies.

\section{Motivation}\label{sec:motivation}

\begin{figure}[h]
  \centering
  \includegraphics[width=1\linewidth]{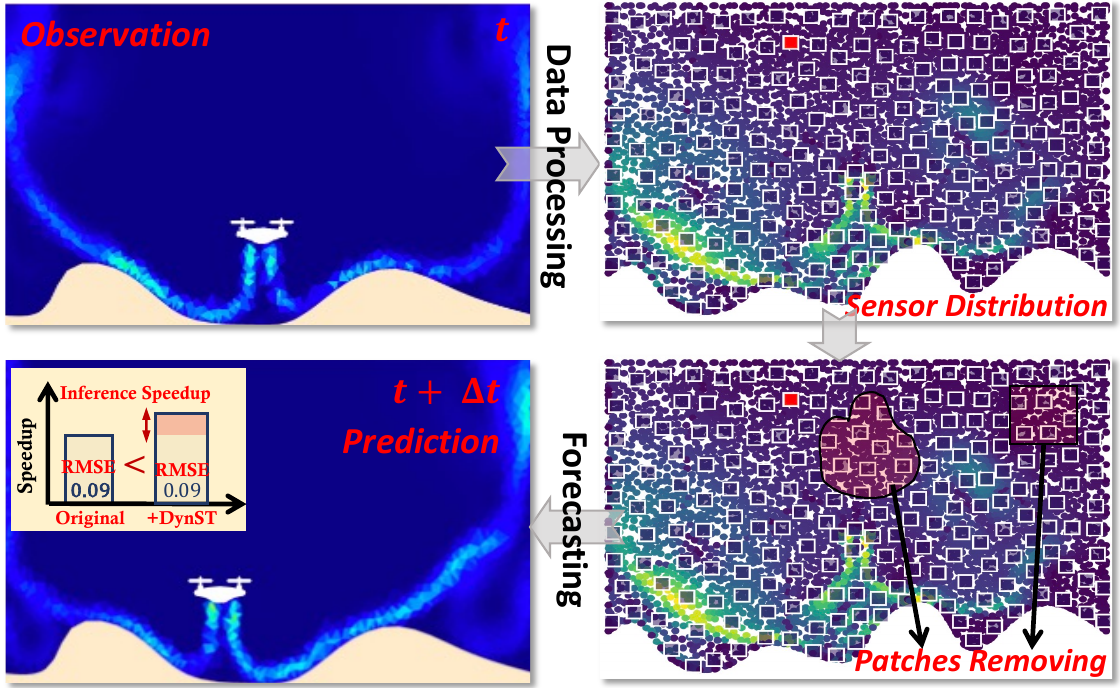}
  \caption{Motivation of our proposal.}
  \label{fig:intro}
\end{figure}

In this section, we carefully examine the significance of our approach and establish the motivation behind DynST. Our analysis begins with empirical observations. Specifically, we use the large-scale dataset EAGLE \cite{janny2023eagle}, designed for learning complex fluid mechanics, as an example. EAGLE is represented as a graph, where each sub-region can be interpreted as the sensory area of a sensor. We demonstrate the important regions using the attention maps from the study and apply masking to the non-essential areas. In each iteration, we randomly mask 15\% of the less important areas to predict the future state of the regions with 7-layer graph convolutional network \cite{kipf2016semi}.

\noindent \textbf{Insights \& Reflections.} As illustrated in Figure 1, we observe that for this dataset, identifying and removing 15\% of the least important patches does not affect the model's performance, which remains consistent with a Root Mean Square Error value about $\sim0.09$. However, the implementation of DynST results in a noticeable speedup in model inference. This finding inspires us to dynamically eliminate non-essential information. By removing these less important regions, we can better identify the parts crucial for future predictions and accelerate inference, which corresponds to sensor deactivation in real-world applications.

\section{Preliminary}

As our research involves both graph and image-type data, we systematically present relevant definitions here to facilitate the demonstration of our model.

\subsection{Graph Notations}
In this study, we focus on an attributed graph, represented as $\mathcal{G} = {{(\mathcal{V}, \mathcal{E})}}$. Here, $\mathcal{V}$ and $\mathcal{E}$ correspond to the node and edge sets, respectively. The graph $\mathcal{G}$ has an associated feature matrix $\mathbf{X} \in \mathbb{R}^{N \times D}$, where $N = |\mathcal{V}|$ indicates the total number of nodes, and $D$ represents the feature dimensionality of each node. For any node $v_i \in \mathcal{V}$, its feature vector is a $D$-dimensional entity $\mathbf{x}_i = \mathbf{X}[i,\cdot]$. The adjacency matrix $\mathbf{A} \in \mathbb{R}^{N \times N}$ defines the inter-node connections, assigning $\mathbf{A}[i,j] = 1$ when a pair of nodes $(v_i, v_j)$ is connected in $\mathcal{E}$ and $0$ otherwise. To effectively learn node representations within $\mathcal{G}$, the majority of GNNs utilize a neighborhood aggregation and message passing paradigm.

\begin{equation}
\mathbf{h}_i^{(l)} = \text{\fontfamily{lmtt}\selectfont \textbf{COMB}}\left( \mathbf{h}_i^{(l-1)}, \text{\fontfamily{lmtt}\selectfont \textbf{AGGR}}\{  \mathbf{h}_j^{(k-1)}: v_j \in \mathcal{N}(v_i) \} \right),\;0\leq l \leq L
\end{equation}
$L$ represents the number of layers in the GNN. The initial feature vector $\mathbf{h}_i^{(0)} = \mathbf{x}_i$ corresponds to the features of node $v_i$. For each layer $l$ in the GNN, where $1\leq l\leq L$, the node embedding of $v_i$ is denoted by $\mathbf{h}_i^{(l)}$. Two critical functions in this process are {\fontfamily{lmtt}\selectfont \textbf{AGGR}} and {\fontfamily{lmtt}\selectfont \textbf{COMB}}. The {\fontfamily{lmtt}\selectfont \textbf{AGGR}} function is responsible for aggregating information from a node’s neighborhood, while the {\fontfamily{lmtt}\selectfont \textbf{COMB}} function is used to combine the representations of the ego-node and its neighbors.

\subsection{Image-type Data Notations}

For effective modeling in image-type data $\mathcal{X}$, we initially divide the total urban area into $p \times p$ sub-regions (patches), with each patch encompassing $(H/p, W/p)$ pixels. $H$ and $W$ are the height and the width of the input images. It is worth noting that the choice of $p$ should balance the trade-off between practicality and spatial granularity. In our implementation, we partition the entire urban area into small squares, each comprising $p \times p$ sensors, adhering to practicality requirements.

\subsection{Problem Formulation}

\label{sec: method}
\begin{figure*}[t]
  \centering
  \includegraphics[width=1\linewidth]{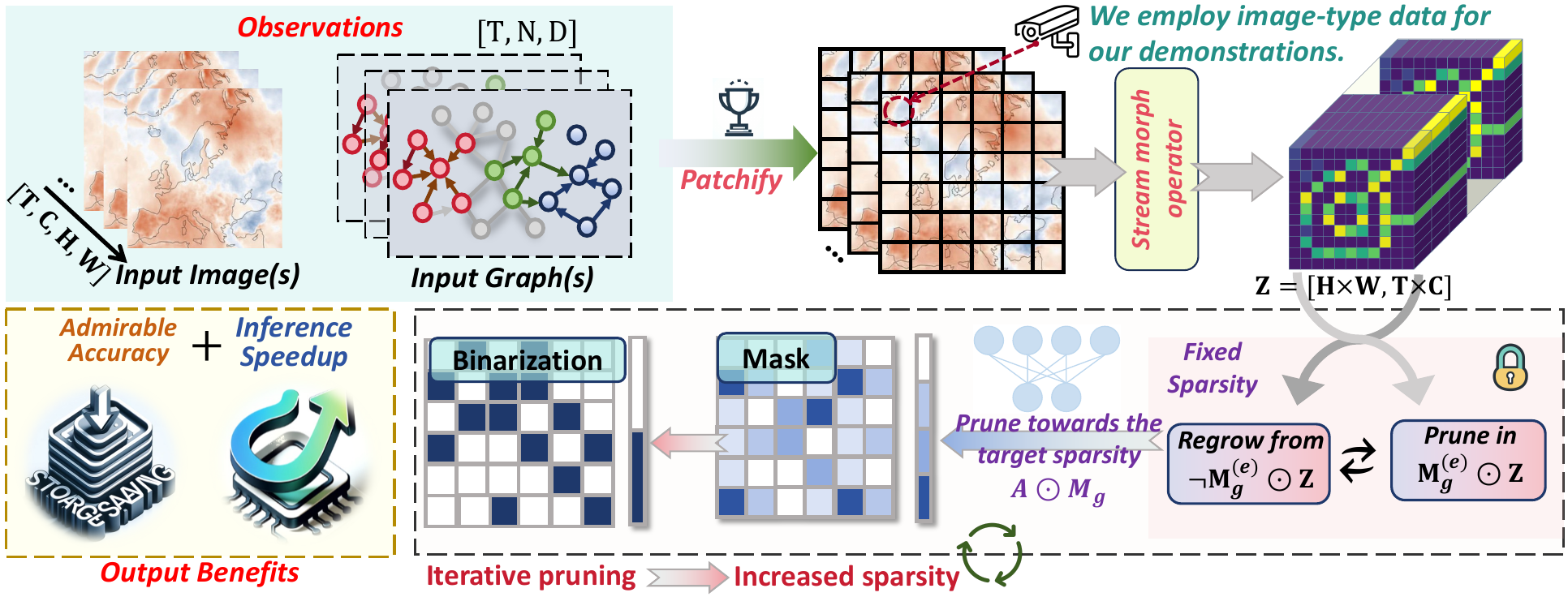}
  \vspace{-0.4cm}
  \caption{Overview of our proposed DynST framework.}
  \label{fig:method}
\end{figure*}

The target of our task is to identify the index of the sparse trivial sub-counterpart of the whole graph ${\mathcal G}$ or image $\mathcal{X}$. For the sake of simplicity in presentation, we eliminate the temporal dimension $T$ from the spatio-temporal data. More formally, we attempt to obtain a trainable mask $M_{e} \in \mathbb{R}^{N}$ (for masking graph nodes) or $M_{e} \in \mathbb{R}^{p \times p}$ (for masking image patches). When we attach $M_{e}$ on original $\mathcal{G}$ ($M_{e}\odot\mathcal{G}$) or on image $\mathcal{X}$ ($M_{e}\odot \mathcal{X}$), the objective is as follows: 
\begin{equation}
\begin{aligned}
&\mathop{\operatorname{maximize}}_{\mathbf{M}_g} \;s_g = 1 - \frac{||\mathbf{M}_g||_0}{||\mathbf{A}||_0}; \;\; {\rm{or}} = 1 - \frac{||\mathbf{M}_g||_0}{p\times p} \\
&\operatorname{s.t.} \left|\mathcal{R}_{DynST}\left( M_{e}\odot*;\Theta \right) - \mathcal{R}_{Ori}(*;\Theta)\right| < \epsilon,
\end{aligned} \label{target}
\end{equation}
where $s_g$ is the sparsity, $||\cdot||_0$ counts the number of non-zero elements, and $\epsilon$ is the threshold for permissible performance difference. $*$ denotes the graph or image inputs and $\mathcal{R}$ represents the evaluation metrics. The above equation expresses that under the sparsity level $s_g$, the model still maintains a loss comparable to that of predictions made with complete data, indicating that there is no significant performance degradation.

\section{Method}

Fig \ref{fig:method} illustrates the overview of the DynST framework. In Earth sciences, sensor deployment typically falls into two categories, \textit{i.e.}, image- and graph-type. Image-type deployment ensures that each area (termed `patch") is well covered by a sensor, while in graph-type deployment, the information from a node can be understood as being collected by a single sensor. To demonstrate the universal capabilities of DynST, we systematically consider both of these deployment types and perform a patchify operation on the images \cite{wu2023solving}. For graph data, since nodes can be defined as sensors, in this study, we do not perform any operations at the data input stage.

\subsection{Stream Morph Operator}

Consider that ST frameworks that receives continuous observation data $\mathcal{X}_i$ at different time steps ($i = 1, 2, ..., T$). According to relevant literature~\cite{arnab2021vivit}, we view this system as a unified four-dimensional structure, \textit{i.e.}, $\mathcal{X}_i \in \mathbb{R}^{[T_{\text{in}}, C_{\text{in}}, H, W]}$. Similarly, the dimensions of a temporal graph can be represented as $\mathcal{G}\in \mathbb{R}^{[T_{\text{in}}, N, D]}$. Typically, in spatio-temporal scenarios, the information collected by sensors is expressed as dynamic temporal observations. \textit{However, while the positions of the sensors are fixed, the sensory data is subject to dynamic changes}. To the best of our knowledge, traditional methods have primarily focused on the optimization of data \cite{anonymous2024nuwadynamics}. We are the first to consider this industrial scenario from the perspective of \textbf{sensor deployment}. As a result, conventional methods are not applicable in our domain. Taking image-type as an example, the image is first tokenized into $N=HW/(p^2)$ non-overlapping patches, then we first introduce the stream morph operator.

\begin{figure}[h]
  \centering
  \includegraphics[width=1\linewidth]{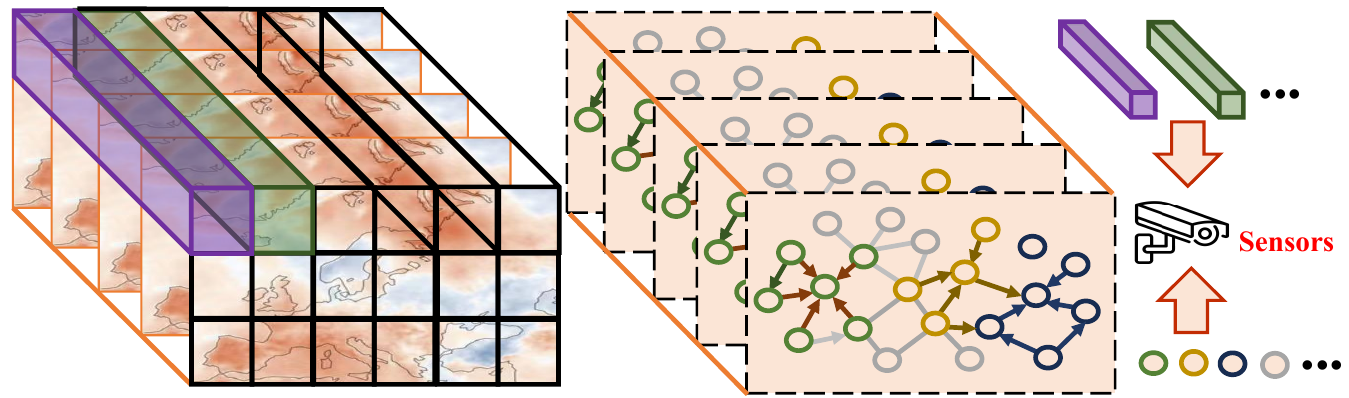}
  \caption{The process of stream morph operator. Each rectangular block and circle node can be interpreted as a sensor recorder.}
  \label{fig:method2}
\end{figure}

As shown in Fig \ref{fig:method2}, stream morph addresses this by merging the $H$ and $W$ channels of the image, and stacking the temporal ($T$) channel with the $C$ channel. This approach effectively eliminates the interference of the $T$ dimension in model predictions. In this way, the training input time series can be deemed as ${\tilde {\mathcal{X}_i}} \in \mathbb{R}^{[H \times W, T_{\text{in}} \times C_{\text{in}}]}$ (graph can be deemed as ${\tilde {\mathcal G}_{in}} \in \mathbb{R}^{[N, T_{\text{in}} \times C_{in}]}$, where $N=HW/(p^2)$), in which each rectangular block ($\tilde {\mathcal X}_{in}^{\left( j \right)} \in \mathbb{R}^{[p^2, T_{\text{in}} \times C_{\text{in}}]}$) and circle node ($\tilde {\mathcal G}_{in}^{\left( j \right)} \in \mathbb{R}^{[1, T_{\text{in}} \times C_{\text{in}}]}$) can be interpreted as a sensor recorder. For ease of understanding, we will primarily use graph inputs as examples to illustrate the model process in subsequent sections. The distinctions between graph-type data and image data will be highlighted in the final \textbf{Model Summary} (Sec \ref{summary}).

Then, stream morph operator employs a parameterized graph mask $M_g \in \mathbb{R}^{[N,1]} $ to dynamically score all nodes, with its parameters shared across all nodes. Given the target graph sparsity $s_g\%$, we first initialize $M_g$ and attach the dense mask $M_g$ on sensor region $M_g \odot {\tilde {\mathcal G}_{in}}$, then we start to resort to currently training scheme to find important and trivial regions.

\subsection{Iterative Pruning towards High Sparsity}

With $M_g$ at hand, we proceed to train the models together with the fixed input graph and the graph mask, denoted as $f(M_g \odot {\tilde {\mathcal G}_{in}},\mathbf{\Theta})$, $f$ denotes the mapping function of the input ST model. with the objective function in Eq.~\ref{target}, we aim to gradually find the sparse sub-graph towards better semantical preservation. One promising approach is to adopt one-shot pruning \cite{ma2021sanity,frankle2020pruning}. However, the sparse mask acquired through one-shot pruning is suboptimal. In fact, the assessment of each sensor necessitates iterative testing to ensure that the removal of a specific area does not significantly impact future predictions. To achieve our objectives, we employ an iterative pruning strategy \cite{chen2021unified} to gradually increase network sparsity. Assuming that each pruning iteration trims $p\%$ of the data parameters, after $\phi$ rounds of pruning, the remaining regions exhibit distinct advantages over the one-shot approach--that is--\textit{By iteratively pruning and retraining, the network can more effectively identify which parts are less important, as the remaining parameters have undergone $\phi$ rounds of repeated verification.} Unlike previous iterative pruning literature, we alternately train the network and the mask $M_g$ to ensure that the mask can fully assimilate the effective information from the training process:

\begin{equation}
    \left. {opt~}_{\Theta}^{(R)}f\left( M_{g}\odot{\overset{\sim}{\mathcal{G}}}_{in},\Theta \right)\leftrightharpoons{opt~}_{M_{g}}^{(M)}f\left( M_{g}\odot{\overset{\sim}{\mathcal{G}}}_{in},\Theta^{*} \right)~ \right.
\end{equation}

\noindent $\leftrightharpoons$ denotes the iterative alternation process. We first train the parameters $\Theta$ for $R$ iterations, then fix $\Theta$ as $\Theta^{*}$ and iteratively train the mask $M_g$ for $M$ iterations. Through this process, the mask $M_g$ potentially encapsulates the important information inherent in the data. Given the target sensor sparsity $s_g\%$, we binarize the mask $M_g$ by zeroing out the parts with the smallest parameter values:

\begin{equation}
    \mathcal{D}o\left( {\rm{ArgTop}} \left(| M_{g}^{(\mu)}| ;p\% \right)\Rightarrow\left\{ 0,1 \right\} \right)
\end{equation}

\noindent $M_{g}^{(\mu)}$ represents the state of the mask $M_{g}$ at the $\mu^{th}$ iteration. The operation ${\rm{ArgTop}}(u,v)$ denotes the process of setting the top $u\%$ parameters in the matrix to 1, while the remaining $v\%$ are set to 0. $\mathcal{D}_{o}$ operator forcefully assigns mask status as 0 or 1.

\subsection{Dynamical Sparse Training}

As depicted above, each sensor region requires meticulous verification to ensure reliability. To this end, in the intervals between each iterative pruning, we further introduce Dynamical Sparse Training (DST) techniques \cite{liu2021we, huang2023dynamic, liu2020dynamic, zhang2023dynamic} to perform fine-tuning between two iterative pruning steps. Concretely, we \textbf{selectively activate a portion of the regions that were previously pruned, while masking the areas that remain unpruned}. After the $\omega^{th}$ round, we perform a \textbf{drop} and \textbf{regrow} process on the pruned mask $M_{g}^{({\omega(R + M)})}$ (\textit{i.e.}, drop \ding{214} regrow). We adjust this process proportion to $q\%$, typically where $q \ll p$, to control the drop and regrow of elements. We perform the ``exchange of sensors" between the current activation regions $\mathcal{E}_{(\omega)} = \mathbf{M}_g \odot {\tilde {\mathcal G}_{in}}$ and its complementary part $\mathcal{E}_{(\omega)}^{C} = \neg\mathbf{M}_g \odot {\tilde {\mathcal G}_{in}}$. Consider that this process at $\omega(D + M)$ time points, we proceed to train and adjust the $M_g$:

\begin{equation}
\begin{aligned}
M_g^{\left( \omega  \right)}\left( {prune} \right) = {\rm{ArgBottom}}\left\{ {\left( {\left| { \nabla \left( {\bar M_g^{\left( \omega  \right)}} \right)} \right|;{\rm{q\% }}} \right) \Rightarrow \left\{ {0,1} \right\}} \right\}
\end{aligned}
\end{equation}

\noindent In this context, ${\bar M_g^{\left( \omega  \right)}}$ represents the elements of $M_g^{\left( \omega  \right)}$ that have not been pruned. Here, we resort to gradient calculation $\nabla$ to identify and drop the elements with the lowest gradients (${\rm{ArgBottom}}$ operator). Generally, gradients can indicate elements with the potential to contribute to the loss function \cite{wang2023brave,evci2020rigging}. We need to align this activation to further explore their effectiveness in future judgments. Going beyond this process, we identify and regrow elements with the highest gradients among those that have been pruned, effectively replacing parts that consist of dropped elements:

\begin{equation}
\begin{aligned}
M_g^{\left( \omega  \right)}\left( {regrow} \right) = \neg {\rm{ArgTop}}\left\{ {\left( {\left| { - \nabla \left( {\neg \bar M_g^{\left( \omega  \right)}} \right)} \right|;{\rm{q\% }}} \right) \Rightarrow \left\{ {0,1} \right\}} \right\}
\end{aligned}
\end{equation}\label{regrow}

\noindent In Eq. \ref{regrow}, we activate elements with larger gradients from the pruned set $\left( {\neg \bar M_g^{\left( \omega \right)}} \right)$. The operation $\neg {\rm{ArgTop}}$ serves as the inverse process of pruning, selecting elements with larger gradients for activation. This ensures that sensor regions with potential contributions are re-evaluated and validated.

Following the completion of the aforementioned evaluation process, we reconstruct $M_g$ to form a more reliable regional mask:

\begin{equation}
\mathbf{M}_g^{(\omega^*)} \leftarrow \left( \mathbf{M}_g^{(\omega)} \setminus M_g^{\left( \omega  \right)}\left( {prune} \right)\right) \cup M_g^{\left( \omega  \right)}\left( {regrow} \right),
\end{equation}
Then, at the beginning of the round $\omega+1$, we continue to train and adjust the mask for sending it to $\omega+1$ round pruning. We binarize the mask $\mathbf{M}_g^{(\omega+1)}$ after another $\Delta T$ iteration training. Without loss of generality, taking the semi-supervised node classification task as an example, our objective function can be expressed as follows:

\begin{equation}
\mathcal{L} (M_g \odot {\tilde {\mathcal G}_{in}}; \Theta) = \frac{1}{K} \sum_{i=1}^K \| \mathcal{Y}_{T+i} - f(M_g \odot {\tilde {\mathcal G}_{in}}; \Theta) \|^2 
\end{equation}

\noindent where $\mathcal{L}$ is the MSE loss calculated over the unmasked node set ${\tilde {\mathcal G}_{in}}$, and $\mathcal{Y}_{T+i}$ denotes the ground-truth.

\begin{figure}[h]
  \centering
  \includegraphics[width=1\linewidth]{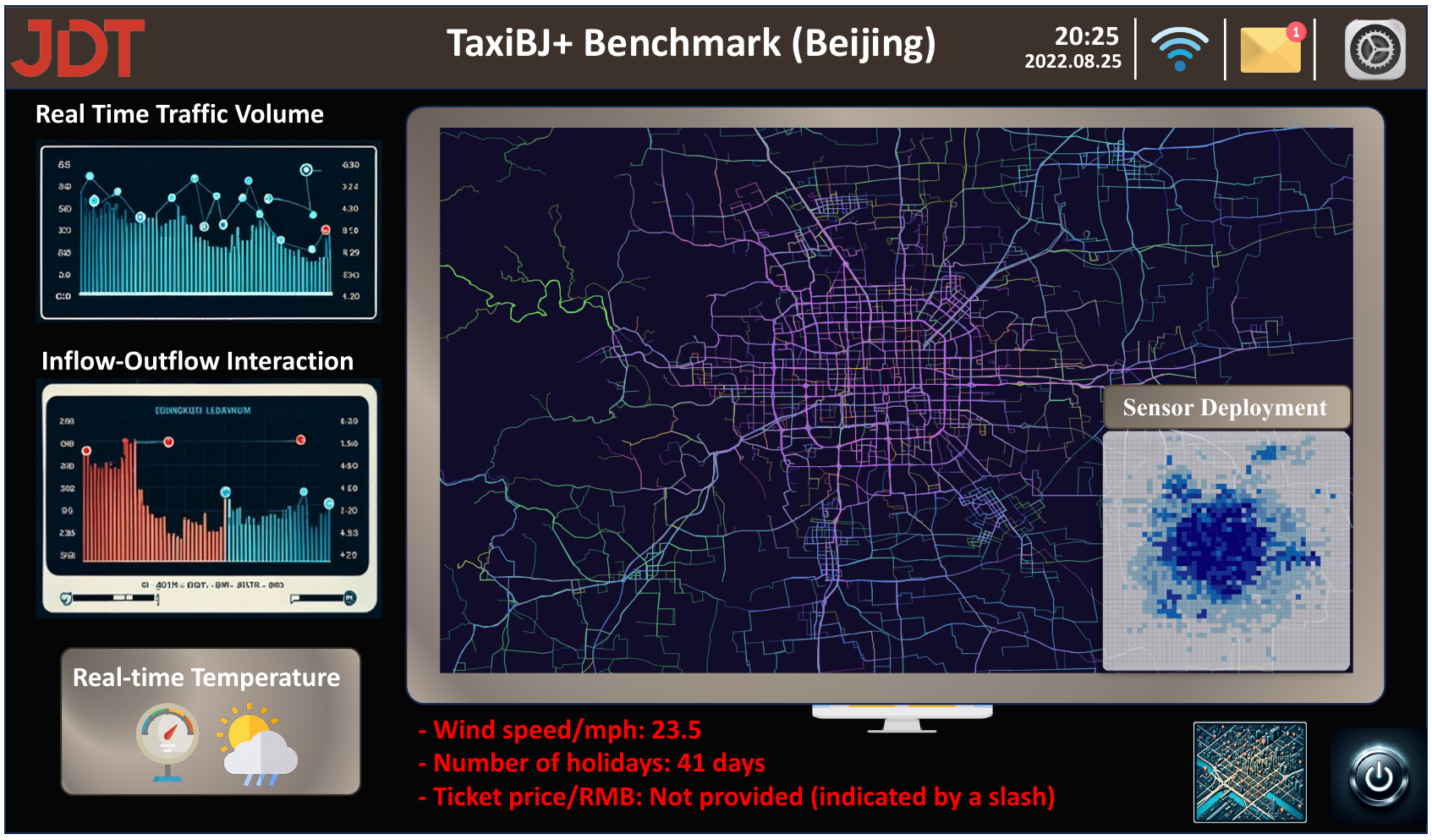}
  \caption{An overview of the anticipated JD Technology Platform, we represent the importance of sensors with a gradient from light to dark blue, effectively removing the deployment in the white areas to emphasize this gradation of significance.}
  \label{fig:intro}
  \vspace{-10pt}
\end{figure}

\subsection{Model Summary \& Complexity Analysis}\label{summary}

For image-type data, we transform each sub-region into a patch, which can also be understood as the concept of a ``node''. Therefore, by training in a similar manner, we can identify the important sub-regions accordingly. DynST can enhance the inference speed of the model, which specifically depends on the predefined sparsity $s_g\%$. Typically, this results in an acceleration ratio of $1/s_g\%$. We summarize our algorithm in Alg.\ref{algorithm}.

\begin{algorithm}
\caption{Dynamic Sparse Training (DynST) Framework}
\begin{algorithmic}[1]
\REQUIRE Input graph $\mathcal{G}_{in}$, Network $f$, Target Sparsity $S_g\%$
\ENSURE Sparse mask $M_g$
\STATE Initialize graph mask $M_g$
\STATE Stream Morph for input $\mathcal{G}_{in} \rightarrow \hat{\mathcal{G}}_{in}$
\WHILE{$1 - \frac{\|M_g\|_0}{\|\mathcal{G}_{in}[;*]\|_0} < S_g$}
\STATE Training network for $R$ iterations
\STATE Training $M_g$ for $M$ iterations
\STATE Dynamical sparse training using Eq.5 and Eq.6
\STATE Adjust $M_g$ using Eq.7
\ENDWHILE
\STATE $\hat{\mathcal{G}}_{in} \leftarrow \hat{\mathcal{G}}_{in} \odot M_g$
\end{algorithmic}\label{algorithm}
\end{algorithm}



\section{Experiments}
In this section, we conduct extensive experiments to answer the following research questions ($\mathcal{RQ}$): 
\begin{enumerate}[start=1,label={\bfseries $\mathcal{RQ}$\arabic*:},leftmargin=3em,itemsep=-0.5mm]
\item Can DynST effectively find the sparse sub-counterpart of the original input without performance degradation?
\item What is the specific performance of DynST on image-type data?
\item What is the specific performance of DynST on graph data?
\item Can we combine the concept of the DynST with a different training scheme?
\end{enumerate}

\noindent To answers these $\mathcal{RQ}$, we orchestrate the following experiments:

\begin{itemize}[leftmargin=*]
    \item \textbf{Main experiment.} We conduct a comprehensive comparative analysis on various scientific datasets, covering meteorology, combustion science, traffic studies, and turbulence dynamics. The study encompasses both mainstream Graph Neural Network (GNN) architectures and non-GNN structures (data pre-processing protocols are placed in appendix~\ref{data_process}).
    
    \item \textbf{Multiple Training Strategies Experiments.} We choose Weatherbench as the benchmark, to evaluate the effectiveness of DynST when combining different training schemes. Specifically, in the training phase, we not only consider the impacts of parallel prediction and autoregressive iterative prediction but also introduce iterative pruning and one-shot pruning strategies.
    
    \item \textbf{Ablation experiment.} We carry out comprehensive ablation studies on the \textbf{Jingdong Technology industry-level traffic} dataset, Taxibj+, to validate the impact of various design choices on the practical implementation of our model. 
\end{itemize}

\noindent \textbf{Experimental settings.} All experiments in this study are conducted on the NVIDIA-A100 40G configuration. To ensure consistency, we use the same settings in all experiments, including learning rate, optimizer, and more. We also apply a uniform training strategy. The loss function used in the experiments is set as Mean Squared Error (MSE) loss. For dataset division, we split the data into training, validation, and test sets in an 8:1:1 ratio. Specifically, for the Vision Transformer model~\cite{ranftl2021vision}, we replace the classification head from the original paper with three deconvolution layers.

    
    


\subsection{Dataset \& Backbones}

\begin{table*}[t]
\caption{Performance comparisons on different GNN and non-GNN architectures, in which we report the best performance of these baselines. All experimental results are under \textbf{\underline{ten runs}}. We show the MAE metric for all settings.}
\label{tab:results1}
\setlength{\tabcolsep}{2.5pt}
\begin{center}
\def\arraystretch{0.94}
\begin{tabular}{ccc|cc|cc|cc|cc|cc|cc|cc} 
\toprule
\multirow{2}{*}{\bf Backbone} & \multicolumn{6}{c}{\bf GNNs}  & \multicolumn{8}{c}{\bf non-GNNs}  & \multirow{2}{*}{\bf Avg Speedup} \\ 
\cmidrule(l){2-7} \cmidrule(l){8-15}
 & \scriptsize \bf STGCN  
 & \scriptsize \bf +  DynST
 & \scriptsize \bf CLCRN  
 & \scriptsize \bf + DynST
 & \scriptsize \bf EGNN 
 & \scriptsize \bf +  DynST 

 & \scriptsize \bf ViT 
 & \scriptsize \bf +  DynST 
 & \scriptsize \bf Simvp 
 & \scriptsize \bf +  DynST
 & \scriptsize \bf TAU
 & \scriptsize \bf +  DynST
  & \scriptsize \bf Earthfarseer
 & \scriptsize \bf +  DynST \\ 
 \midrule
   \multicolumn{10}{l}{\scriptsize{  \demph{ \it{Model Performance Evaluation} }} }\\
    WeatherBench \ding{168} &4.35  &4.37  &1.17  &1.22  &2.98  &3.00  &0.72  &0.73  &0.74  &0.73  & 0.73 & 0.77 &0.58  &0.62 & 1.721\\
    WeatherBench \ding{169} &2.02  &2.04  &1.49  &1.52  &3.39  &3.42  &0.27  &0.29  & 0.27 & 0.29 &0.26  & 0.25 &0.24  & 0.25& 1.522\\
    WeatherBench \ding{170} &0.79  &0.75  &0.45  &0.47  & 0.66  &0.72  &0.24  & 0.26 &0.25  & 0.26 &0.23  & 0.24 &0.22  & 0.24& 1.119\\
    WeatherBench \ding{171} &3.64  &3.67 & 1.33 &1.31  & 2.31 &2.33  &0.51  & 0.54 & 0.51 & 0.52 & 0.49 & 0.50 & 0.48 &0.50 &1.398 \\
    FIT $\phi$ &1.27  &1.29  &0.97  &0.98  & 1.03& 1.09 &0.23  &0.22  &0.14  & 0.16 &0.13  &0.14  &  0.09& 0.11&1.543\\
    FIT $\varphi$ &0.96  &1.09  &0.76  &0.81  & 0.92  &0.95  & 0.17 & 0.19 &0.10  & 0.09 &0.09  & 0.10 &0.02  & 0.03&1.541 \\
    Taxibj+ Inflow &5.98  &5.99  &3.98  &4.02  & 4.22 & 4.33 & 3.22 &3.33  &3.05  &3.11  &2.98  & 3.00 & 2.09 & 2.10 &1.421\\
    Taxibj+ Outflow &5.21  &5.23  &3.64  &3.60  & 4.21 & 4.19 &3.67  & 3.59 & 3.01 &3.03  &2.77  & 2.87 & 2.12 &2.22 & 1.987\\
    EAGLE &1.99  &2.07  &1.45  &1.47  & 1.66 & 1.67 &1.45  &1.47  &1.23  &1.34  &1.19  &1.27  &1.08  &1.12& 1.988 \\
    \midrule
\end{tabular}
\label{main1}
\end{center}
\end{table*}

\begin{table}[h] 
\footnotesize
\setlength{\tabcolsep}{2.5pt}
  \caption{Comparison results among different benchmarks, considering different data sparsity levels and prediction lengths.} 
  \centering
  \scalebox{1}{
  \begin{tabular}{cccccccc}
    \toprule
     \multirow{2}{*}{\makecell{\textbf{Benchmark}\\Graph-type}} &  \multicolumn{3}{c}{\textbf{Taxibj+}}  &  \multicolumn{3}{c}{\textbf{EAGLE}}  \\
     
     \cmidrule(lr){2-4} \cmidrule{5-7} 
     
    & \textbf{4} &  \textbf{8}  &  \textbf{12} & \textbf{30} &  \textbf{40}  &  \textbf{50}  \\
    \midrule
    $10\%$   & ${1.92_{\pm0.01}}$& ${1.99_{\pm0.01}}$ & ${2.03_{\pm0.01}}$  &  ${1.14_{\pm0.02}}$ &${1.18_{\pm0.02}}$ & ${1.19_{\pm0.02}}$ \\
    
    $20\%$  & ${2.04_{\pm0.03}}$  &${2.12_{\pm0.01}}$& ${2.14_{\pm0.01}}$ & ${1.17_{\pm0.02}}$ & ${1.23_{\pm0.02}}$& ${1.24_{\pm0.02}}$ \\
      
    $30\%$   &${2.07_{\pm0.01}}$  &${2.17_{\pm0.01}}$ & ${2.18_{\pm0.02}}$& ${1.21_{\pm0.02}}$& ${1.24_{\pm0.01}}$ & ${1.26_{\pm0.02}}$   \\

    $40\%$   & ${2.21_{\pm0.02}}$  & ${2.24_{\pm0.01}}$ & ${2.25_{\pm0.01}}$ &${1.25_{\pm0.03}}$ & ${1.26_{\pm0.01}}$ & ${1.28_{\pm0.02}}$ \\

    $50\%$ & ${2.37_{\pm0.03}}$   & ${2.39_{\pm0.01}}$ & ${2.42_{\pm0.03}}$ & ${1.27_{\pm0.01}}$ & ${1.27_{\pm0.02}}$ & ${1.29_{\pm0.02}}$ \\
    
    $60\%$ &${2.45_{\pm0.02}}$  & ${2.48_{\pm0.01}}$ & ${2.51_{\pm0.01}}$ & ${1.29_{\pm0.01}}$ & ${1.30_{\pm0.01}}$ & ${1.33_{\pm0.02}}$ \\

    \midrule
    \multirow{2}{*}{\makecell{\textbf{Benchmark}\\Image-type}}  &  \multicolumn{3}{c}{\textbf{FIT $\phi$}}  &  \multicolumn{3}{c}{\textbf{WeatherBench \ding{168}}}  \\
     
     \cmidrule(lr){2-4} \cmidrule{5-7} 
     
    & \textbf{30} &  \textbf{40}  &  \textbf{50} & \textbf{4} &  \textbf{8}  &  \textbf{12}  \\
    \midrule
    $10\%$   & ${0.13_{\pm0.01}}$ & ${0.15_{\pm0.01}}$ & ${0.15_{\pm0.01}}$   & ${0.53_{\pm0.01}}$ & ${0.57_{\pm0.01}}$ & ${0.58_{\pm0.02}}$ \\
    
    $20\%$  & ${0.15_{\pm0.03}}$  &${0.16_{\pm0.01}}$ & ${0.17_{\pm0.01}}$ & ${0.54_{\pm0.01}}$ & ${0.59_{\pm0.01}}$ & ${0.61_{\pm0.03}}$\\
      
    $30\%$   & ${0.15_{\pm0.01}}$  & ${0.17_{\pm0.01}}$ & ${0.18_{\pm0.01}}$  & ${0.58_{\pm0.01}}$ & ${0.62_{\pm0.01}}$ & ${0.64_{\pm0.01}}$  \\

    $40\%$   & ${0.17_{\pm0.02}}$  & ${0.18_{\pm0.03}}$ & ${0.19_{\pm0.01}}$  & ${0.60_{\pm0.01}}$ & ${0.65_{\pm0.01}}$ & ${0.66_{\pm0.02}}$ \\

    $50\%$ & ${0.19_{\pm0.01}}$  & ${0.21_{\pm0.01}}$ &${0.22_{\pm0.02}}$  & ${0.61_{\pm0.01}}$ & ${0.67_{\pm0.01}}$ & ${0.69_{\pm0.01}}$ \\
    
    $60\%$ & ${0.21_{\pm0.01}}$  & ${0.22_{\pm0.01}}$ &${0.23_{\pm0.02}}$  & ${0.63_{\pm0.01}}$ & ${0.69_{\pm0.01}}$ & ${0.70_{\pm0.01}}$ 
   \\
    \bottomrule
  \end{tabular}\label{tab:rq2}
  }
\end{table}
\begin{figure}[h]
  \centering
  \includegraphics[width=1\linewidth]{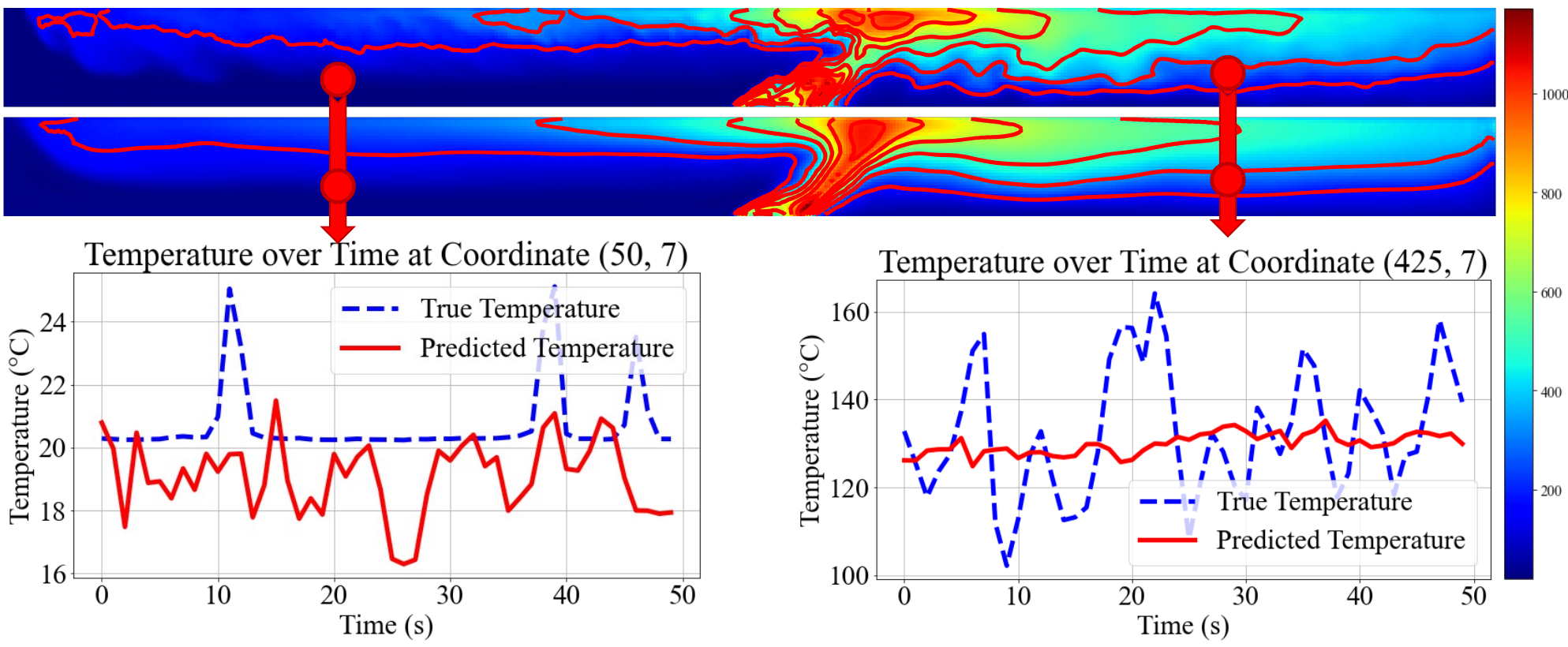}
  \caption{The performance visualization of FIT datasets. We can see that the overall temperature deviation is within 10 degrees Celsius, meeting the requirements of the fire science field.~\cite{emmons1986needed}}
  \label{fig:Fire_wendu}
\end{figure}

\begin{figure*}[htbp]
  \centering
  \includegraphics[width=0.98\linewidth]{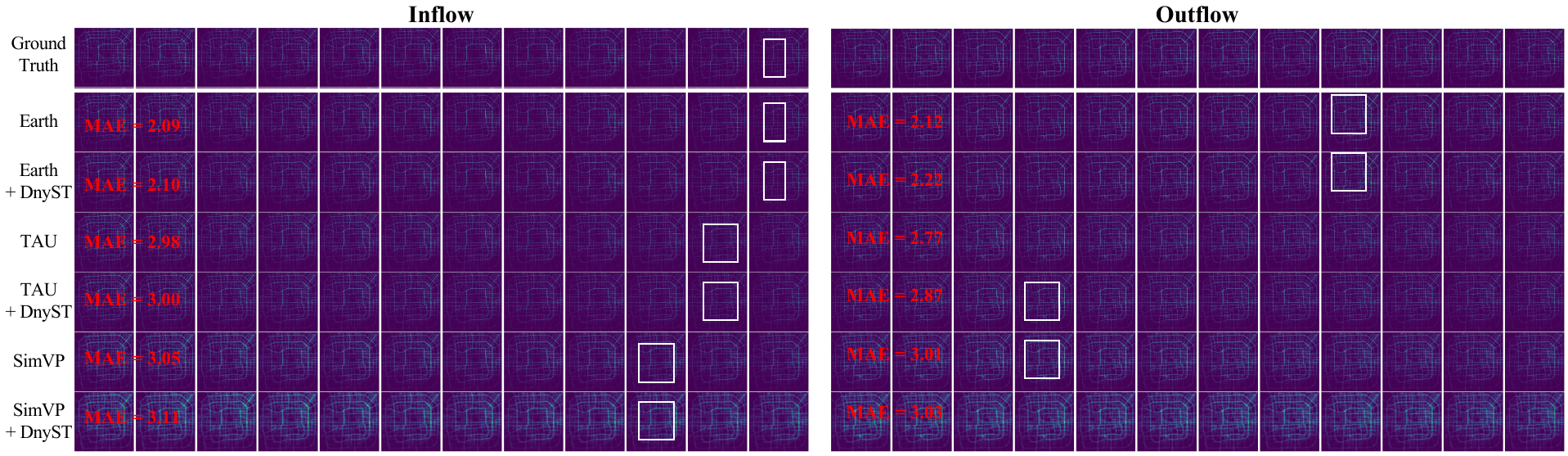}
  \caption{The performance visualization of JD Technology Taxibj+ datasets. It displays the traffic inflow and outflow prediction results of various methods such as Earthfarseer, TAU, SimVP, and their versions combined with DnyST, along with their MAE scores. The left side shows the comparison of the Ground Truth with the predictions for each method, and the right side presents the outflow predictions.}
  \label{fig:Taxibj}
\end{figure*}

\noindent \textbf{Datasets.} In this study, we conduct thorough analyses of multiple sensor-loaded datasets covering four main areas: meteorology, fires, turbulence, and traffic flow. In meteorology, we select the Weatherbench dataset. Following the design framework of related papers~\cite{rasp2020weatherbench}, we consider four key variables: temperature (\ding{168}), humidity (\ding{169}), wind speed (\ding{170}), and cloud cover (\ding{171}), with the dataset containing 2048 nodes. For fire data, we choose the FIT dataset. Adhering to existing paper settings~\cite{anonymous2023spatiotemporal}, we focus primarily on two variables: temperature ($\phi$) and visibility ($\varphi$), totaling 15360 data nodes. In turbulence, we refer to the EAGLE dataset~\cite{janny2023eagle}, a large turbulence dataset involving velocity and pressure variables, presented in an irregular grid form with 162760 nodes. Regarding traffic flow, we use JD Technology's Taxibj+ dataset~\cite{wu2023earthfarseer}, which provides traffic flow statistics for Beijing city, comprising 16384 data nodes.~\textbf{For the convenience of this study, each node is considered an independent sensor.}

\noindent \textbf{Backbones.} We use both GNN and non-GNN architectures to systematically validate the generalizability of our ideas. Concretely, we use GNN-based models as our backbone, such as STGCN~\cite{han2020stgcn}, CLCRN~\cite{lin2022conditional} and EGNN~\cite{satorras2021n}, as well as non-GNNs such as Vision Transformer~\cite{dosovitskiy2021an}, SimVP-V2~\cite{tan2022simvp}, TAU~\cite{tan2023temporal} and Earthfarseer~\cite{wu2023earthfarseer}. All GNNs have 7-layer encoder blocks, while non-GNNs use Transpose Conv2d for upsampling. This detailed categorization method greatly helps in deeply understanding and accurately analyzing the capabilities of DynST.


\subsection{Main experiments ($\mathcal{RQ}$1)}

In this section, we test whether DynST can effectively remove non-essential areas (corresponding to the concept of sensors in the real world) without impacting the overall predictive performance of the model. To thoroughly investigate the generalizability and optimization capabilities of DynST, we integrate it with existing general frameworks and set the iterative pruning process to occur 10 times, each time reducing the data by 3\%. We showcase the main results in Tab \ref{main1} and we can list the observations:

\textbf{Obs 1.DynST has demonstrated that the removal of certain parts from the input data does not affect the model's performance.} As shown in Tab~\ref{tab:results1},  We can easily observe the outcomes following the integration of the {\fontfamily{lmtt}\selectfont \textbf{DynST}} concept into the model (+DynST). In the GNN architecture, the addition of DynST generally has a minimal impact on MAE. For example, on the WeatherBench \ding{168} and FIT $\varphi$ datasets, the MAE slightly increases from 4.35 to 4.37 and from 0.92 to 0.95, respectively. In non-GNN architectures, The addition of DynST still has a minimal impact on the MAE. For instance, in the ViT architecture on the Taxibj+ Outflow dataset, the MAE decreases from 3.67$\rightarrow$3.59. In particular, DynST generally significantly enhances the inference speed across various architectures. For example, in WeatherBench \ding{168}, STGCN speeds up to 1.721 times, EGNN on FIT $\varphi$ to 1.541 times, and ViT on Taxibj+ Outflow to 1.987 times.

\textbf{Obs 2. DynST shows high efficiency in several scenarios.} DynST is also highly effective in improving the inference efficiency of various architectures. For example, on the WeatherBench \ding{168} dataset, the inference speed of STGCN increased by 23.7\% with DynST (from the original speed to 1.721 times faster). Similarly, on the FIT $\varphi$ dataset, the EGNN architecture achieved a 14.5\% speed increase with DynST (reaching 1.541 times faster). Moreover, on the Taxibj+ Outflow dataset, the inference speed of the ViT architecture almost doubled, specifically a 34.5\% increase (rising to 1.987 times faster). These examples collectively show DynST's capability to significantly enhance computational efficiency in various scenarios. The percentage-based speed improvements highlight its notable advantage in accelerating the inference of various ST architectures.

\subsection{Deep insights ($\mathcal{RQ}$2 \& $\mathcal{RQ}$3)}

In this section, we conduct a more systematic study of DynST's ability to accelerate inference. We select both graph and image-type data to observe model performance at various levels of sparsity. Concretely, for graph-type data, we choose Taxibj+ and  EAGLE as benchmarks. For image-type data, we choose temperature ($\phi$) variable of FIT datasets and the temperature (\ding{168}) variable of the WeatherBench as verification. We integrate it with existing general frameworks and set the iterative pruning process to occur 10 times, with each iteration reducing the data volume by $\{1\%,2\%,\cdots,6\%\}$. Then we can obtain the data sparsity $\{10\%,20\%,\cdots,60\%\}$. We employ roll out strategy \cite{luo2023hope} to iteratively predict long sequence and verify the long-term prediction ability of baselines after involving DynST. We list the observations as follows.

\textbf{Obs 3. DynST effectively achieves long-term predictions without causing significant performance degradation.}  We tested the capability of long-term prediction with a combination of DynST and Earthfarseer and found that incorporating the concept of dynamic sparse training did not compromise the model's performance. Even at a higher sparsity level of 60\%, it still manages to deliver reasonably good predictive performance without a significant increase in RMSE.

\textbf{Obs 4. DynST effectively meets industrial-level requirements (30\% sparsity), helping to achieve manageable inference demands while reducing the burden of inference.}  As shown in Fig \ref{fig:Fire_wendu}, The first line of the display meticulously captures the actual observed temperature flow field, providing a vivid and accurate representation of the existing conditions.  In contrast, the second line offers a predictive perspective, showcasing the temperature flow field as forecasted by the innovative Earthfarseer+DynST model.  This juxtaposition not only illustrates the capabilities of the predictive model but also allows for a direct comparison between observed and predicted states. \textbf{Bottom}: Delving deeper into the analysis, the left image opens a window into a detailed time series comparison. It meticulously charts both the real and the predicted temperatures at the specific coordinates of (50,7), offering a granular view of the model's precision over time. Similarly, the right image extends this comparison to another set of coordinates, (425,7), revealing how the model captures the temporal evolution of temperatures in this distinct area. These results showcase the remarkable ability of the DynST-enhanced model to preserve high local fidelity. This fidelity is not just theoretical; it translates into practical, industry-level reliability, consistently maintaining the prediction deviation within a tight 15\% margin \cite{verda2021expanding}. Such performance not only underscores the robustness of the Earthfarseer+DynST model but also highlights its potential for widespread application in scenarios demanding high precision and reliability (Fig \ref{fig:Taxibj} also supports our research findings).

\subsection{Structural \& Ablation study ($\mathcal{RQ}$4)}

We initially configure DynST to maintain the model at a moderate sparsity level (30\%) to observe how well the model preserves structural integrity at this level of sparsity. Here, we employ two metrics, SSIM and PSNR, to measure the completeness of the model's predictions. Higher values of SSIM and PSNR indicate more accurate structural predictions by the model. Additionally, we also observe the trend of SSIM performance at different levels of sparsity.

\begin{table}[h]
\footnotesize
    \setlength{\abovecaptionskip}{2pt}
    \setlength{\belowcaptionskip}{2pt}
    \centering 
    \setlength{\tabcolsep}{8pt} 
    \caption{SSIM and PSNR results on three research domains. The underline symbol represents the best performance. Ori denotes the original results, +Dyn denotes add DynST at sparsity 30\%.} 
    \resizebox{0.48\textwidth}{!}{ 
    \begin{tabular}{c|c|c}
          \hline
    Model (data) & SSIM (Ori $\leftrightarrow$ +Dyn) & PSNR (Ori $\leftrightarrow$ +Dyn)  \\ \hline \hline
    SimVP (TaxiBJ+)         & $0.94_{ \pm 0.01 }$/ $0.93_{ \pm 0.01 }$   & $36.27_{ \pm 0.01 }$/ $35.43_{ \pm 0.01 }$       \\
    TAU (TaxiBJ+)         &$0.96_{ \pm 0.01 }$/ $0.95_{ \pm 0.01 }$     & $36.76_{ \pm 0.01 }$ / $35.62_{ \pm 0.01 }$             \\
    Earthfarseer (TaxiBJ+)      &  $\underline{0.98}$ $_{ \pm 0.01 }$/ $\underline{0.96}$ $_{ \pm 0.01 }$        & $\underline{37.84}$ $_{ \pm 0.01 }$/ $\underline{36.44}$ $_{ \pm 0.01 }$      \\ \hline \hline
    
    CLCRN (WeatherBench)    & $0.94_{ \pm 0.02 }$/ $0.93_{ \pm 0.02 }$   & $36.12_{ \pm 0.02 }$/ $35.22_{ \pm 0.19 }$        \\
    Simvp (WeatherBench)    & $0.96_{ \pm 0.01 }$ / $0.95_{ \pm 0.01 }$            & $37.33_{ \pm 0.01 }$/ $36.33_{ \pm 0.17 }$           \\
    Earthfarseer (WeatherBench)      & $\underline{0.98}$$_{ \pm 0.01 } $/ $\underline{0.97}$ $_{ \pm 0.01 } $       & $\underline{39.27}$ $_{ \pm 0.11}$/ $\underline{38.12}$ $_{ \pm 0.03}$     \\ \hline \hline

    VIT (FIT)         & $0.90_{ \pm 0.02 }$/ $0.89_{ \pm 0.02 }$     & $35.41_{ \pm 0.02 }$/ $33.33_{ \pm 0.01 }$       \\
    EGNN (FIT)         & $0.83_{ \pm 0.01 }$/ $0.81_{ \pm 0.01 }$        & $35.41_{ \pm 0.01}$/ $34.68_{ \pm 0.02}$  \\
    Earthfarseer (FIT)      & $\underline{0.95}$ $_{ \pm 0.01 }$ / $\underline{0.93}$ $_{ \pm 0.01 }$     & $\underline{37.23}$ $_{ \pm 0.01 }$/ $\underline{36.31}$ $_{ \pm 0.01 }$       \\ \hline
    \end{tabular}} \label{tab:ssim}
\end{table}

\begin{figure}[h]
  \centering
  \includegraphics[width=1\linewidth]{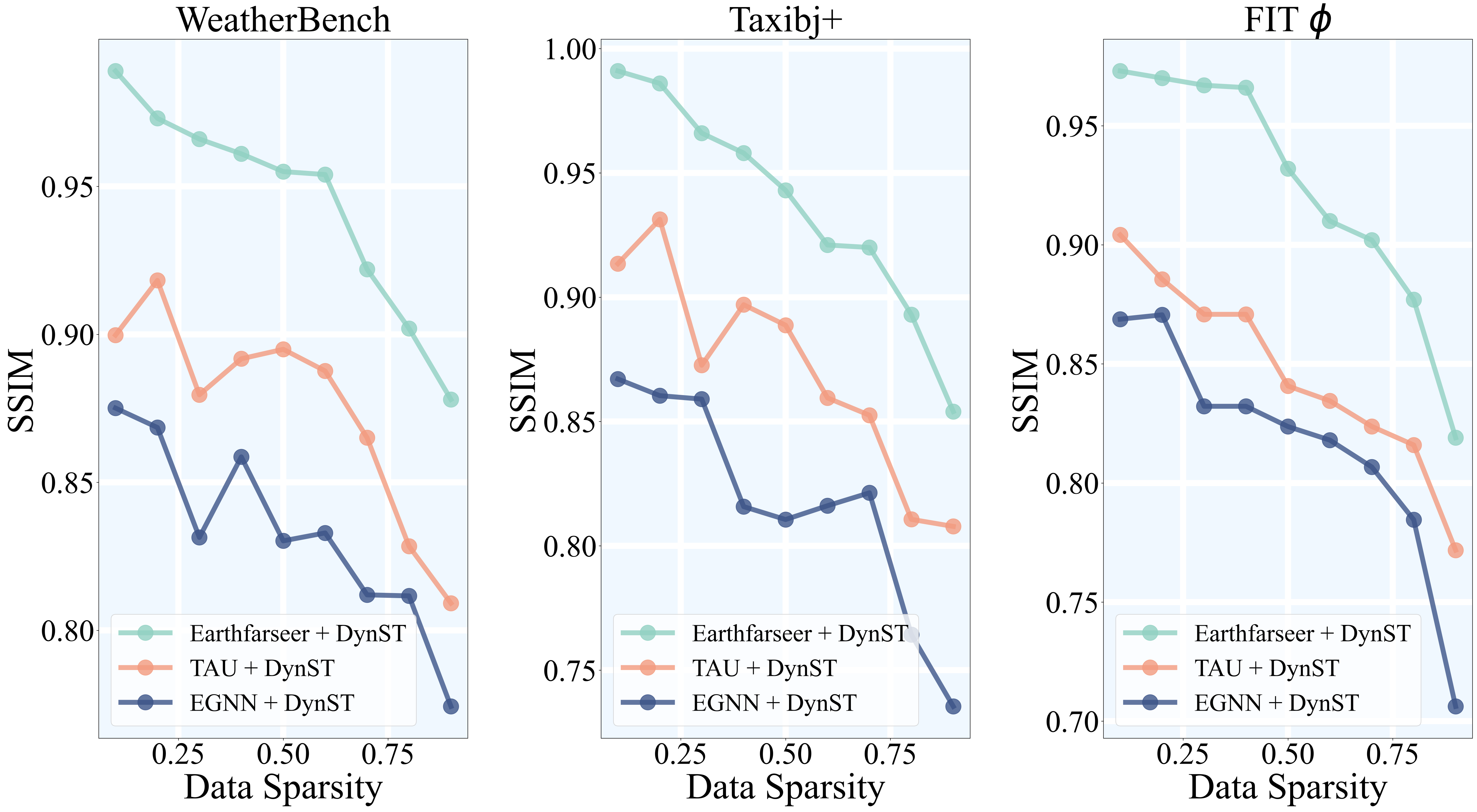}
  \caption{The proposed three plug-and-play model + DynST on SSIM.}
  \label{fig:ssim}
\end{figure}

\textbf{Obs 5.} As shown in Tab \ref{tab:ssim} and Fig \ref{fig:ssim}, we find that integrating DynST into the model does not significantly impact the SSIM and PSNR metrics. On the TaxiBJ+ dataset, Earthfarseer achieves an SSIM value close to 0.97, and the incorporation of DynST appears to have minimal effect on the prediction results. This phenomenon is nearly identical on both the WeatherBench ($0.98\rightarrow0.97$) and FIT ($0.98\rightarrow0.93$) datasets, thereby validating the effectiveness of DynST. Further, as the model's SSIM values under varying data sparsity levels (Fig \ref{fig:ssim}), we note that as sparsity increases, the SSIM values gradually decrease, providing a trade-off solution for practical applications.

In the last, we select three training schemes (Earthfarseer as the base model) to explore the performance of our algorithm and the benefits of combining our algorithm with mainstream training approaches: (1) \textbf{One-shot pruning (OP)}: We thoroughly train our model and subsequently conduct training of the mask for a one-time pruning process. (2) \textbf{Iterative pruning (IP)}: As our work can be regarded as a pruning method, we have opted for the widely recognized iterative pruning (IP) strategy \citep{frankle2018lottery} in the main manuscript part, we prune data for 10 times and every time for pruning 4\% sub-counterpart. (3) \textbf{Dynamic sparse training (DST)}: We select a target sparsity level and then maintain the data training consistently at this fixed sparsity. Dynamically, we remove and restore the smallest and largest magnitudes in the mask \cite{evci2020rigging}. We set a 40\% sparsity for dynamic training. Table \ref{tab:trainscheme} shows STGCN, SimVP, and Earthfarseer's performance in IP, OS, and DST training methods. Their RMSEs are 0.5698, 0.5108, 0.3507 (IP), 0.6197, 0.5650, 0.4121 (OS), and 0.5792, 0.5261, 0.3495 (DST). In summary, our DST performs the best.

\begin{table}[h]
  \centering
  \setlength{\tabcolsep}{4.0pt}
  \caption{Performance across different training schemes (RMSE).}
\begin{tabular}{c|c|ccc}
    \toprule
    & \multirow{2}{*}{\tabincell{c}{Baselines}} & \multicolumn{3}{c}{Training Schemes} \\
    & & IP & OS & DST \\
    \midrule
    \tabincell{c}{EAGLE} & \tabincell{c}{STGCN \\ SimVP \\ Earthfarseer} & \tabincell{c}{0.5698 \\ 0.5108 \\ 0.3507} & \tabincell{c}{ 0.6197 \\0.5650 \\ 0.4121} & \tabincell{c}{0.5792 \\ 0.5261 \\ 0.3495} \\
    \midrule
\tabincell{c}{FIT $\phi$} & \tabincell{c}{STGCN \\ SimVP \\ Earthfarseer} & \tabincell{c}{0.3245 \\ 0.2193 \\ 0.1983} & \tabincell{c}{0.3617 \\ 0.2565 \\ 0.2293} & \tabincell{c}{0.3123 \\ 0.2252 \\ 0.1842} \\
    \bottomrule
  \end{tabular}\label{tab:trainscheme}
\end{table}

\section{Conclusion}

In this paper, we introduce the concept of dynamic sparse training in the context of sensor deployment, \textbf{termed DynST}, which adjusts sensor deployment dynamically through training without compromising the model's predictive capabilities. DynST ingeniously circumvents the complexity issues posed by the temporal dimension through clever dimension mapping. Following this, through dynamic training and mask operations, we can precisely identify the less significant parts of the output data, which correspond to the areas detected by the sensors.



\begin{acks}
This work is mainly supported by the National Natural Science Foundation of China (No. 62402414). This work is also supported by Tencent (CCF-Tencent Open Fund, Tencent Rhino-Bird Focused Research Program), Didi (CCF-DiDi GAIA Collaborative Research Funds), Guangzhou Municipal Science and Technology Project (No. 2023A03J0011) and Guangzhou-HKUST(GZ) Joint Funding Program (No. 2024A03J0620).
\end{acks}

\bibliographystyle{ACM-Reference-Format}
\balance 
\bibliography{main}

\clearpage
\appendix 
\section{Datasets and backbones descriptions.} 
\label{data_backbones}

\begin{table}[h]
\centering
\caption{The statistics of the datasets.}
\label{tab:dataset_statistics}
\begin{tabular}{lcccc}
\toprule
Dataset & \#Nodes & \#Variables & \#Input & \#Output  \\
\midrule
Weatherbench & 2048 &4& 12 & 12  \\
FIT & 15360 &2& 50 & 50  \\
Taxibj+ &16384& 2& 12 & 12  \\
EAGLE & 3388 &2& 50& 50\\
\bottomrule
\end{tabular}
\end{table}

In this study, we analyze four benchmark datasets. Each snapshot in these datasets serves as an independent graph structure. We summarize the statistical characteristics of these datasets in Table \ref{tab:dataset_statistics}. Specifically, the datasets include:

\noindent \textbf{1. Weatherbench Dataset:} Each graph contains 2048 nodes, covering four variables: temperature, humidity, wind speed, and cloud concentration. The input and output duration for this dataset is 12 time steps.

\noindent \textbf{2. FIT Dataset:} Each graph in this dataset consists of 15360 nodes, with two variables: temperature and visibility. The input and output duration is 50 time steps.

\noindent \textbf{3. Taxibj+ Dataset:} Each graph has 16384 nodes, including two variables: Inflow and Outflow. The input and output duration is 12 time steps.

\noindent \textbf{4. EAGLE Dataset:} Each graph in this dataset comprises 3388 nodes, with two variables: pressure and speed. The input and output duration is 50 time steps.

These datasets provide diverse experimental scenarios and analytical perspectives for our research.

\section{Dataset preprocessing.} 
\label{data_process}

\begin{figure}[h]
  \centering
  \includegraphics[width=1\linewidth]{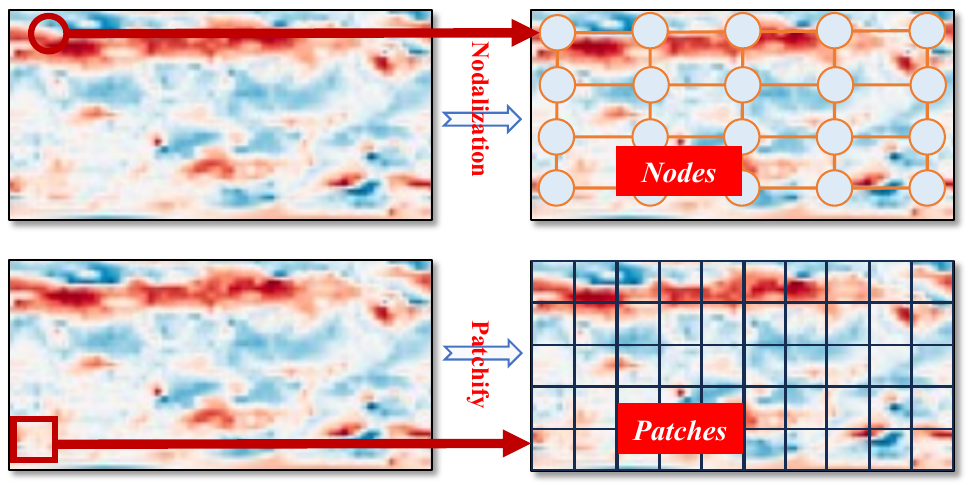}
  \caption{Transforming Raw Data into Graph and Image Structures.}
  \label{fig:dataprocess}
\end{figure}

In this section, we meticulously detail the specifics of data processing, as shown in Figure~\ref{fig:dataprocess}, encompassing the conversion of raw data into graph and image formats through two distinct processes: Nodalization and Patchify. We utilize Weatherbench as a case study to illustrate these concepts:

\noindent \textbf{1. Nodalization:} This process involves the dimensional transformation of raw data from the format $(C, H, W)$, where $C$ represents the number of physical variables, and $H$ and $W$ signify the data's height and width, respectively. In this context, the data can be perceived as having $H \times W$ observation points, each containing $C$ variables. If we analogize each observation point to a sensor, these correspond to nodes in a graph structure. Consequently, the transformed graph data dimension is $(Num\_nodes, C)$, where $Num\_nodes = H \times W$. To alleviate memory pressure during training, a down-sampling of $H$ and $W$ can be implemented in practical applications.

\noindent \textbf{2. Patchify:} In the Patchify process, we adhere to the strategy outlined in the literature, assuming that each Patch is of size $p \times p$. This results in a total of $(H/p) \times (W/p)$ Patches. The dimension of each Patch is $(p \times p \times C)$. This method enables us to leverage Transformer-based architectures for data feature extraction. At the same time, for convolutional structures, the raw data can be directly inputted into the model without the need for specialized data preprocessing.

Through these two methodologies, we effectively transform the original data format into one that is conducive to deep learning model processing, thereby enhancing the efficiency of data handling and model training.
\end{document}